\documentclass[letterpaper]{article} 
\usepackage{aaai23}  
\usepackage{times}  
\usepackage{helvet}  
\usepackage{courier}  
\usepackage[hyphens]{url}  
\usepackage{graphicx} 
\usepackage{natbib}  
\usepackage{caption} 
\usepackage{newfloat}
\usepackage{listings}
\usepackage{amsthm}
\usepackage{booktabs}
\usepackage{latexsym}
\usepackage[utf8]{inputenc}
\usepackage{multirow}
\usepackage{pifont}
\usepackage{bm}
\usepackage{amssymb}
\makeatother

\usepackage[linesnumbered,ruled,vlined]{algorithm2e}
\usepackage{algpseudocode}  
\usepackage{amsmath}  
\title{Logic and Commonsense-Guided Temporal Knowledge Graph Completion}
\author{
    Guanglin Niu\textsuperscript{\rm 1},
    Bo Li\textsuperscript{\rm 1,2}\thanks{\ \ Corresponding author.}
}
\affiliations{
\textsuperscript{\rm 1} Institute of Artificial Intelligence, Beihang University, Beijing, China\\
\textsuperscript{\rm 2} Hangzhou Innovation Institute, Beihang University, Hangzhou, China\\
\{beihangngl, boli\}@buaa.edu.cn
}

\usepackage{bibentry}

\begin{document}

\maketitle

\begin{abstract}
A temporal knowledge graph (TKG) stores the events derived from the data involving time. Predicting events is extremely challenging due to the time-sensitive property of events. Besides, the previous TKG completion (TKGC) approaches cannot represent both the timeliness and the causality properties of events, simultaneously. To address these challenges, we propose a \textbf{L}ogic and \textbf{C}ommonsense-\textbf{G}uided \textbf{E}mbedding model (LCGE) to jointly learn the time-sensitive representation involving timeliness and causality of events, together with the time-independent representation of events from the perspective of commonsense. Specifically, we design a temporal rule learning algorithm to construct a rule-guided predicate embedding regularization strategy for learning the causality among events. Furthermore, we could accurately evaluate the plausibility of events via auxiliary commonsense knowledge. The experimental results of TKGC task illustrate the significant performance improvements of our model compared with the existing approaches. More interestingly, our model is able to provide the explainability of the predicted results in the view of causal inference. The source code and datasets of this paper are available at \url{https://github.com/ngl567/LCGE}.
\end{abstract}

\section{Introduction}

\begin{figure}
    \centering
    \includegraphics[scale=0.37]{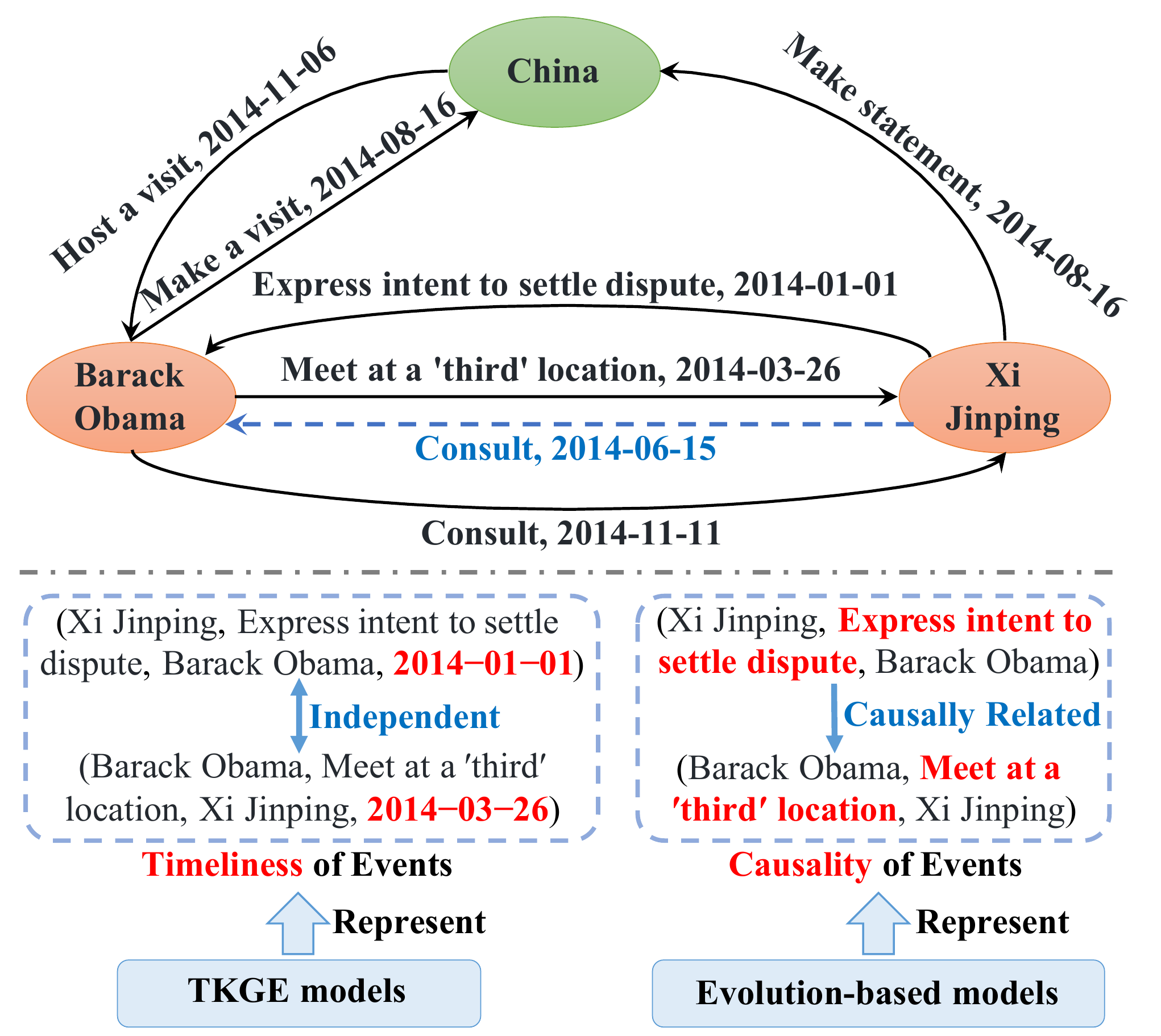}
    \caption{A brief example of the TKG from ICEWS dataset and the TKGC task predicting the missing event $(Xi\ Jingping, Consult, Barack\ Obama, 2014-06-15)$. Besides, the existing TKGC models cannot jointly represent the timeliness of events and the causality among events.}
    \label{fig:intro}
\end{figure}

Knowledge graph (KG) has been developed rapidly in recent years, which stores facts in the form of (subject, predicate, object). To further exploit the events involving time, temporal KG (TKG) represents each event as a quadruple (subject, predicate, object, time) where the time information can be formulated by a timestamp or time interval. For instance, an event $(Barack\ Obama,$ $Consult, Xi\ Jinping,$ 2014-11-11$)$ in a TKG as shown in Figure~\ref{fig:intro} indicates that this event occurs on the definite date of 2014-11-11.

Temporal KG completion (TKGC) is an essential technique to predict whether some potential events missing in the TKG will occur, i.e., $(Xi\ Jingping, Consult, $ $Barack\ Obama,\  $2014-06-15$)$ shown in Figure~\ref{fig:intro}. Particularly, an event is only valid at a specific time namely the timeliness. The existing TKGC approaches can be classified into two categories: (1) the evolution-based models are capable of representing the causality among events to reason the future events, such as Know-Evolve~\cite{know-evolve} RE-NET~\cite{REnet} and CyGNet~\cite{zhu-etal-2021-cygnet}. As the declaration of causality of events is shown in Figure~\ref{fig:intro}, when two events occur in certain time order, one event has an impact on the other. The event occurring earlier is the reason and the event occurring later is the result. (2) The TKG embedding (TKGE) models, which this paper focuses on, evaluate the plausibility of events via scoring events with embeddings of entities and predicates together with timestamps, including TTransE~\cite{TTransE}, HyTE~\cite{HyTE} and DE-SimplE~\cite{DE-SimplE}. TKGE models regard the events that occur at different times are completely independent and these approaches predict the events at the known time.

However, the previous models face several challenges: (1) the existing TKGC models believe that any TKG simply contains events involving time but they ignore the long-term effective commonsense knowledge implied in the TKG. (2) The evolution-based models struggle to reason about events with weak correlations to past events while the TKGE models are unable to exploit the causality among events. In summary, all the existing TKGC approaches cannot jointly represent the timeliness and causality of events. (3) Almost all the previous TKGC techniques are data-driven without explainability. Besides, StreamLearner~\cite{StreamLearner} is the only known approach that automatically mines temporal rules from TKGs. However, it merely explores the single pattern that all the atoms in the rule body are restricted at the same time but ignores the diverse temporal rule patterns among events.

To address the above challenges, we develop a novel and effective \underline{\textbf{L}}ogic and \underline{\textbf{C}}ommonsense-\underline{\textbf{G}}uided \underline{\textbf{E}}mbedding (\textbf{LCGE}) model to represent events more adequately for improving the performance of TKGC. Concretely, we design a temporal rule-guided predicate embedding regularization for learning the causality property of events. Furthermore, a joint event and commonsense-based KG embedding strategy is proposed to score each event via learning the time-sensitive representation involving timeliness and causality as well as the time-independent representation in the view of commonsense. The main contributions of our work include:

\begin{itemize}
    \item We design a temporal rule-guided regularization strategy to inject the causality among events into predicate embeddings. To the best of our knowledge, it is the first effort to introduce temporal rules into TKGE models.
    
    \item We model each event from the perspectives of both the time-sensitive representation and the commonsense, facilitating higher accuracy in predicting missing events.
    
    \item The experimental results on three benchmark datasets of TKGs illustrate the significant performance improvements of our model compared with several state-of-the-art baselines. More interestingly, our model could provide explainability via temporal rules.
\end{itemize}

\begin{figure*}
    \centering
    \includegraphics[scale=0.46]{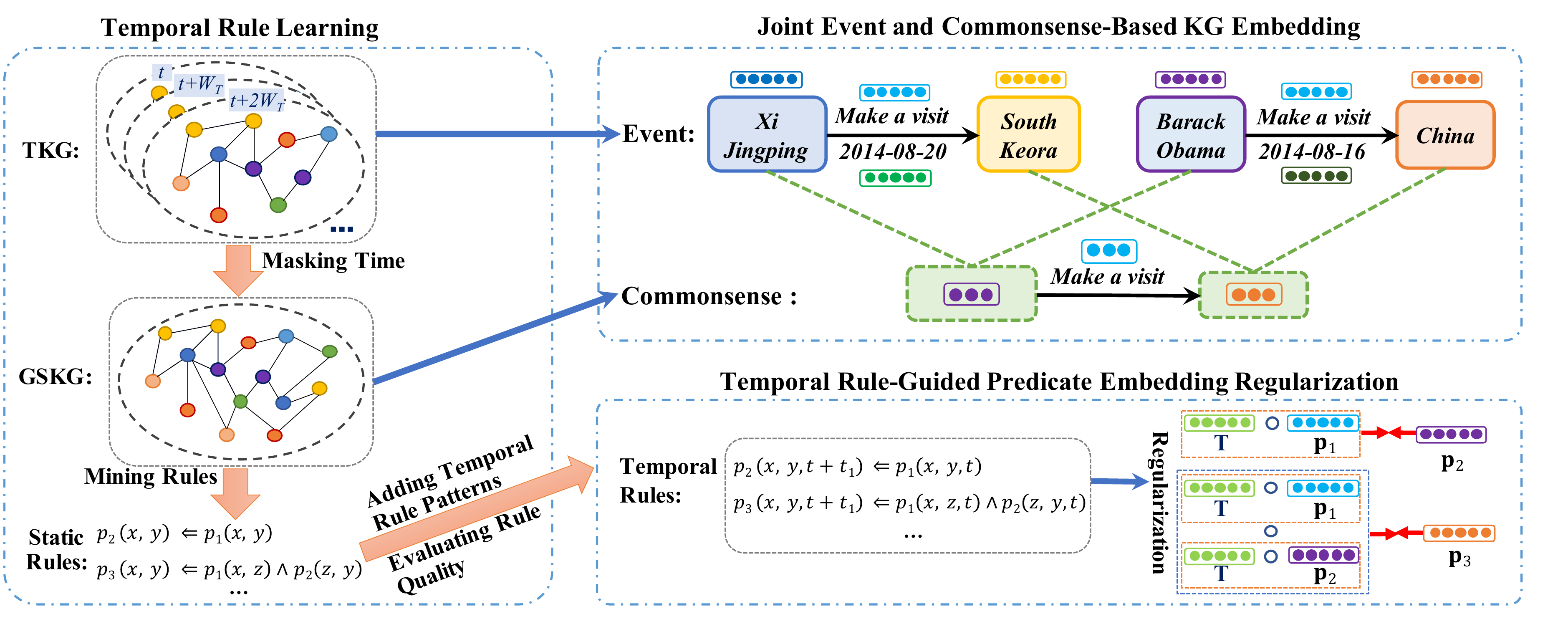}
    \caption{The whole framework of our LCGE model. Commonsense is represented as the interaction between two entity concept embeddings and a predicate embedding. The subject concept embeddings of the two events shown in this figure should be close to each other since they are associated with the same predicate. In the temporal rule-guided predicate embedding regularization module, $\textbf{T}$ denotes the temporal transformation operator that will be described in section~\ref{sec:3.3}.}
    \label{fig:framework}
\end{figure*}

\section{Related Work}

\subsection{Traditional KGE Models}
KGE technique aims to score the plausibility of facts via learning the entity and predicate embeddings. TransE~\cite{Bordes:TransE} models the interaction among a triple fact via regarding a predicate as the translation operation from subject to object. More advanced approaches represent predicates as rotation operations for modeling the symmetric and anti-symmetric predicates, such as RotatE~\cite{RotatE}, QuatE~\cite{QuatE} and DualE~\cite{DualE}. RESCAL~\cite{RESCAL} conducts three-order tensor decomposition to calculate the truth values of facts. DistMult~\cite{Distmult} and ComplEx~\cite{Trouillon:ComplEx} simplify RESCAL model with fewer parameters and improve the performance of KG completion.

\subsection{Temporal KGE Models}
Temporal KGE (TKGE) models extend the traditional KGE approaches by supplementing the time embeddings to represent the time-aware events. TTransE~\cite{TTransE} adds an extra time embedding in the translation-based score function to represent the timeliness of events. Motivated by the hyper-plane specific to the relation proposed in TransH~\cite{Wang:TransH}, HyTE~\cite{HyTE} projects the entities and predicates into the hyper-plane of a specific time. DE-SimplE~\cite{DE-SimplE} leverages the diachronic entity embeddings to represent each entity in different timestamps. ATiSE~\cite{ATiSE} learns the time-aware embeddings of entities and predicates to a Gaussian distribution for representing the time uncertainty. TeRo~\cite{TERO} extends HyTE to learn the time-sensitive entity and predicate embeddings via rotation operation specific to various timestamps~\cite{RotatE}. TComplEx~\cite{TComplEx} upgrades ComplEx to score each event via a fourth-order tensor decomposition that introduces time information.

\subsection{Rule Learning}
According to the symbolic characteristics of KGs, logic rules are naturally suitable for KG completion task. The Horn rule is a typical type of logic rule in the form of $a_1 \Leftarrow a_2 \wedge a_3 \wedge \cdots \wedge a_n$, in which $a_1$ denotes an atom in the rule head (namely head atom) and $a_2, \cdots a_n$ are the atoms in the rule body (namely body atoms). Some effective rule learning algorithms are developed specifically for large-scale KGs relying on rule searching and rule quality evaluation, such as AMIE+~\cite{Galarrage:AMIE}, ScaLeKB~\cite{ScaLeKB}, RuLES~\cite{RuLES}, Anyburl~\cite{AnyBurl}, DRUM~\cite{DRUM}, RLvLR~\cite{RLvLR} and RNNLogic~\cite{RNNLogic}. However, these approaches are designed for static KGs rather than TKGs. StreamLearner~\cite{StreamLearner} is the only known algorithm to mine the temporal rules of which all the body atoms are restricted, simultaneously.

\section{The Proposed LCGE Model}
In this section, we firstly introduce the preliminaries (§\ref{sec:3.1}). Then, we declare our developed temporal rule learning algorithm with diverse temporal rule patterns (§\ref{sec:3.2}) and further propose the temporal rule-guided predicate embedding regularization (RGPR) (§\ref{sec:3.3}). Afterward, the joint event and commonsense-based KGE mechanism (§\ref{sec:3.4}) together with the overall optimization objective (§\ref{sec:3.5}) are presented. The whole framework of our model is shown in Figure~\ref{fig:framework}.

\subsection{Preliminaries}
\label{sec:3.1}
\textbf{Temporal Knowledge Graph}. The temporal knowledge graph (TKG) is a set of events attached with time information. Each event is represented as a quadruple $(s, p, o, t)$, in which $s$ and $o$ are subject and object, $p$ denotes the predicate, and $t$ implies the timestamp or time interval. Particularly, an event with a time interval $[t_s, t_e]$ can be converted into two events with timestamps namely $(s, p, o, t_s)$ and $(s, p, o, t_e)$.

\noindent \textbf{Temporal Rule}. A temporal rule is formulated as the conjunction of the atoms attached with time labels. In our work, we focus on the temporal rules in the form of
\begin{equation}\small
    p_{n+1}(x, y, t) \Leftarrow p_1(x, z_1, t_1) \wedge \cdots \wedge p_n(z_{n-1}, y, t_n)
\end{equation}
where $p_i(i=1,\cdots,n+1)$ are the predicates. $x$, $y$ and $z_j(j=1,\cdots,n-1)$ denote the entity variables. Particularly, $t$ and $t_l(l=1,\cdots,n)$ indicate the time variables that satisfy the constraint $t_1 \leq t_2 \leq \cdots \leq t_n \leq t$. A temporal
rule signifies that the rule head will happen if the rule body holds. A grounding of the temporal rule is obtained by replacing the variables with specific entities and timestamps.

\subsection{Temporal Rule Learning}
\label{sec:3.2}
We develop a novel static-to-dynamic strategy to mine the temporal rules with various patterns. In the static rule learning stage, we first convert all the quadruple events in training set into triples via masking the time information of each event as shown in Figure~\ref{fig:framework}. Then, we obtain a global static KG (GSKG) consisting of all the triples and mine the static rules from the GSKG by any existing rule learning algorithm such as AMIE+~\cite{Galarrage:AMIE} or Anyburl~\cite{AnyBurl}. Notably, temporal rules can be regarded as the extension of static rules with various temporal rule patterns.

\begin{figure*}
    \centering
    \includegraphics[scale=0.42]{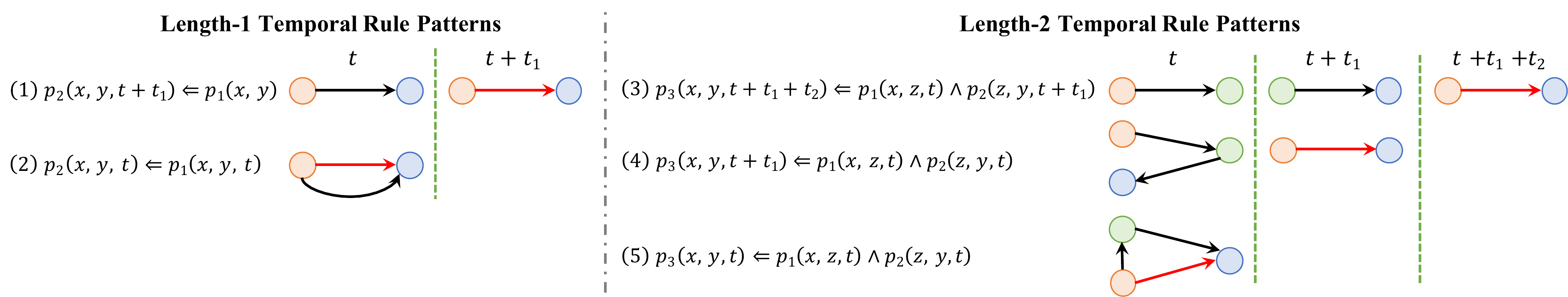}
    \caption{A brief diagram of the proposed five temporal rule patterns. The dots indicate the variables. The red lines denote the predicates in the head atoms while the black lines represent the predicates in the body atoms.}
    \label{fig:trule_pattern}
\end{figure*}

\subsubsection{Formulation of Temporal Rule Patterns.}
At the dynamic rule learning stage, five temporal rule patterns as shown in Figure~\ref{fig:trule_pattern} are well-designed according to the diverse temporal sequences among atoms as followings:


(1) A length-1 rule in which the two atoms have different timestamps: $p_2(x, y, t+t_1) \Leftarrow p_1(x, y, t)$.

(2) A length-1 rule where the two atoms are valid at the same time: $p_2(x, y, t) \Leftarrow p_1(x, y, t)$.

(3) A length-2 rule that the timestamps of the three atoms are different from each other: $p_3(x, y, t+t_1+t_2) \Leftarrow p_1(x, z, t) \wedge p_2(z, y, t+t_1)$.

(4) A length-2 rule where the timestamps of the body atoms differ from that of the head atom: $p_3(x, y, t+t_1) \Leftarrow p_1(x, z, t) \wedge p_2(z, y, t)$.

(5) A length-2 rule for the three atoms to be valid at the same time: $p_3(x, y, t) \Leftarrow p_1(x, z, t) \wedge p_2(z, y, t)$.

A detailed description of temporal rule patterns can be found in Appendix A.1.

\subsubsection{Quality Evaluation of Temporal Rules.}
On account of the static rules mined by the existing rule learning algorithms and the proposed five temporal rule patterns, we extend each static rule to be the corresponding candidate temporal rules according to the defined temporal rule patterns.

To evaluate the quality of each candidate temporal rule, we firstly merge the events in the same time window into a sub-graph since some causally related events may occur in the same sub-graph or the adjacent sub-graphs. Then, we search for all the events satisfying the grounding of candidate temporal rules. Specific to a length-1 temporal rule in the general form $p_2(x, y, t+T_1) \Leftarrow p_1(x, y, t)$, it satisfies pattern (1) when $T_1=0$ and signifies pattern (2) if $T_1>0$. Thereby, following the definition of the evaluation criteria in previous rule learning models~\cite{DRUM}, we propose support degree (SD), standard confidence (SC), and head coverage (HC) of temporal rules as follows:
\begin{align}
    &SD=
    \begin{cases}
    \#(a,b):p_1(a,b,t)\land p_2(a,b,t), \ \ T_1=0 \\
    \#\left(a,b\right):\sum_{t_{1i}\in\left[1,W_T\right]}{p_1\left(a,b,t\right)\land p_2(a,}\\
    {\quad \quad \quad \ \ \ \ \ b, t+t_{1i})},\qquad \qquad \quad T_1>0 \label{eq1} 
    \end{cases}\\
    &SC=
    \begin{cases}
    \frac{SD}{\#\left(a,b\right):p_1\left(a,b,t\right)}, \qquad \qquad \qquad \ \ \ \ \ \ T_1=0\\
    \frac{1}{\vert W_T \vert }\cdot\frac{SD}{\#\left(a,b\right):p_1\left(a,b,t\right)},\qquad \qquad \ \ \ T_1>0  \label{eq2}
    \end{cases}\\
    &HC=
    \begin{cases}
    \frac{SD}{\#\left(a,b\right):p_2\left(a,b,t\right)}, \qquad \qquad \qquad  \ \ \ \ \ T_1=0\\
    \frac{SD}{\#\left(a,b\right):\sum_{t_{1i}\in\left[{1,W}_T\right]}{p_2\left(a,b,t+t_{1i}\right)}},\ \ \ \ T_1>0 \label{eq3}
    \end{cases}
\end{align}
in which $W_T$ denotes the size of the pre-defined time window. The evaluation criteria of length-2 temporal rules are defined similarly to that of length-1 temporal rules and are provided in Appendix A.2.

We acquire the SC and HC of each candidate temporal rule by traversing the timestamp $t$ in accordance with the temporal rules' evaluation criteria. The candidate temporal rules satisfying the thresholds of SC and HC are employed for the following regularization strategy in section~\ref{sec:3.3}. The entire algorithm of our proposed temporal rule learning module is presented in Appendix A.3.

\subsection{The Designed RGPR Mechanism}
\label{sec:3.3}
Based on the generated temporal rules signifying the causality among atoms, we define a time transfer operator $\textbf{T}$, which ensures all the atoms in a rule are represented at the same time to calculate their correlations. Take the temporal rule $p_\mathrm{3}\left(x,\ y,\ {\ t+t_1+t}_2\right)\ \Leftarrow{\mathrm{\ }p}_\mathrm{1}\left(x,\ z,\ t\right)\land p_\mathrm{2}\left(z,\ y,\ \ t+t_1\right)$ for an instance, projecting the body atom $p_1\left(x,\ z,\ t\right)$ into the same time window of the head atom requires time transfer operation twice. Furthermore, the temporal rule-guided predicate embedding regularization $\textbf{G}$ is proposed to inject the causality implied in the temporal rules into predicate embeddings corresponding to each temporal rule pattern:

(1) $p_2(x, y, t+t_1) \Leftarrow p_1(x, y, t)$:
\begin{equation}
    \mathbf{G}= \Vert \left(\mathbf{T}\circ \mathbf{p}_{r1}\right)-\mathbf{p}_{r2} \Vert 
    \label{eq7}
\end{equation}

(2) $p_2(x, y, t) \Leftarrow p_1(x, y, t)$:
\begin{equation}
    \mathbf{G}= \Vert \mathbf{p}_{r1}-\mathbf{p}_{r2} \Vert 
    \label{eq8}
\end{equation}

(3) $p_3(x, y, t+t_1+t_2) \Leftarrow p_1(x, e, t) \wedge p_2(e, y, t+t_1)$:
\begin{equation}
    \mathbf{G}= \Vert \left(\mathbf{T}\circ\mathbf{T}\circ \mathbf{p}_{r1}\right)\circ\left(\mathbf{T}\circ \mathbf{p}_{r2}\right)-\mathbf{p}_{r3} \Vert 
    \label{eq9}
\end{equation}

(4) $p_3(x, y, t+t_1) \Leftarrow p_1(x, e, t) \wedge p_2(e, y, t)$:
\begin{equation}
    \mathbf{G}= \Vert \left(\mathbf{T}\circ \mathbf{p}_{r1}\right)\circ\left(\mathbf{T}\circ \mathbf{p}_{r2}\right)-\mathbf{p}_{r3} \Vert 
    \label{eq10}
\end{equation}

(5) $p_3(x, y, t) \Leftarrow p_1(x, e, t) \wedge p_2(e, y, t)$:
\begin{equation}
    \mathbf{G}= \Vert \mathbf{p}_{r1}\circ \mathbf{p}_{r2}-\mathbf{p}_{r3} \Vert 
    \label{eq11}
\end{equation}
where $\mathbf{p}_{r1}$, $\mathbf{p}_{r2}$ and $\mathbf{p}_{r3}$ indicate the embeddings of predicates $p_1$, $p_2$ and $p_3$, respectively. $\circ$ denotes Hardmard product. Particularly, we theoretically prove that the designed $\mathbf{G}$ could model the causality among events in Appendix A.4.

\subsection{Joint Event and Commonsense-Based KGE}
\label{sec:3.4}
We take full advantage of the long-term validity of commonsense to evaluate the plausibility of an event accurately since some events that go against commonsense will never happen. Thus, we model each event from the perspective of both time-sensitive and time-independent representations.

To learn the time-sensitive representation of events, inspired by TKGE model TComplEx~\cite{TComplEx}, we learn the timeliness of each event embodied with the timestamp via four-order tensor decomposition. Besides, the causality among events can be represented via our RGPR mechanism together with the subject and object embeddings. Given an event quadruple $(s, p, o, t)$, the time-sensitive score function is defined as
\begin{align}
    E_1&\left(s,\ p,\ o,\ t\right)=\mathrm{Re}\left(\mathbf{s}^\top\ diag(\mathbf{p}_{t} + {\mathbf{p}}_{r})\ \bar{\mathbf{o}}\right) \nonumber \\
    &=\mathrm{Re}\left(\sum_{i=1}^{d}{\left[\mathbf{s}\right]_i\cdot\left[\mathbf{p}\circ\mathbf{t}+\mathbf{p}_{r}\right]_i\cdot\left[\bar{\mathbf{o}}\right]_i}\right)
    \label{eq12}
\end{align}
where $\mathbf{s} \in \mathbb{C}^d$, $\mathbf{p} \in \mathbb{C}^d$, $\mathbf{o} \in \mathbb{C}^d$ indicate the embeddings in the $d$-dimension complex vector space with regard to $s$, $p$ and $o$, respectively. $\bar{\mathbf{o}}$ is the conjugate of $\mathbf{o}$. Particularly, $\mathbf{p}_t=\bf{p}\circ\bf{t}$ denotes the predicate embedding constrained by the timestamp $t$. Meanwhile, $\mathbf{p}_r$ is the causality representation of the predicate $p$ learned by our RGPR mechanism. From Eqs. \ref{eq7}-\ref{eq11}, if some events hold, the other events that are causally relevant to these events would have higher scores according to the regularization of predicate embeddings. $\left[\mathbf{x}\right]_i$ denotes the $i$-th value of the complex vector $\mathbf{x}$. Based on the developed score function in Eq.~\ref{eq12}, we could jointly represent the timeliness and the causality of events, facilitating a more sufficient time-sensitive representation of events.

\begin{table*}
 \centering
 \renewcommand\tabcolsep{12.0pt}
 \begin{tabular}{c|cc|ccc|c}
 \toprule
Dataset		& \#Predicate	& \#Entity    & \#Train	& \#Valid	& \#Test    & Time Span \\
 \midrule
 ICEWS14	& 230	& 6,869    & 72,826     & 8,941	    & 8,963     & 2014\\
 ICEWS05-15 & 251	& 10,094   & 368,962	& 46,275	& 46,092    & 2005-2015\\
 Wikidata12k & 24	& 12,554   & 32,497	    & 4,062	    & 4,062     & 1479-2018\\
 \bottomrule
 \end{tabular}
 \caption{Statistics of the experimental datasets. Time span indicates the range of years in which the events occur.}
 \label{table1}
 \end{table*}
 
\begin{table*}\small
 \centering
 \renewcommand\tabcolsep{12.0pt}
 \begin{tabular}{l|l}
 \toprule
Temporal Rule Patterns		& The Mined Temporal Rules with Confidence\\
 \midrule
 $\emph{p}_2\left(x, y, t+t_1\right)\ \Leftarrow p_{1}\left(x,\ y,\ t\right)$	& $\emph{Host\ a\ visit}\left(x,\ y,\ t+t_1\right)\ \Leftarrow Make\ a\ visit\left(x,\ y,\ t\right)\ 0.45$ \\
 \midrule
 $\emph{p}_{2}\left(x,\ y,\ t\right)\ \Leftarrow p_{1}\left(x,\ y,\ t\right)$    & $\emph{Mobilize\ or\ increase\ armed\ forces}\left(x,\ y,\ t\right)\ \Leftarrow Increase\ \emph{military\ alert\ status}\left(x,\ y,\ t\right)\ 0.33$ \\
 \midrule
 $\emph{p}_{3}\left(x,\ y,\ t+t_1+t_2\right) \Leftarrow$ & $\emph{Express\ intent\ to\ meet\ or\ negotiate}\left(x,\ y,\ t+t_1+t_2\right)\Leftarrow\ $ $\emph{Engage\ in\ negotiation}^{-1}\left(x,\ z,\ t\right) $ \\ $\emph{p}_{1}\left(x,\ z,\ t\right)\land p_{2}\left(z,\ y,\ t+t_1\right)$ &  $\land\ Express\ intent\ to\ \emph{cooperate}\left(z,\ y,\ t+t_1\right)\ 0.27$ \\
  \midrule
  $\emph{p}_{3}\left(x,\ y,\ t+t_1\right)\ \Leftarrow$   & $\emph{Protest\ violently,\ riot}\left(x,\ y,\ t+t_1\right)\Leftarrow Demonstrate\ or\ \emph{rally}\left(x,\ y,\ t\right)\land Make\ an\ appeal$\\
  $\emph{p}_{1}\left(x,\ y,\ t\right)\land p_{2}\left(x,\ y,\ t\right)$  & $or\ request\left(x,\ y,\ t\right)\ 0.25$\\
  \midrule
  $\emph{p}_3\left(x,\ y,\ t\right)\ \Leftarrow p_1\left(x,\ z,\ t\right)$ & $\emph{Engage\ in\ negotiation}\left(x,\ y,\ t\right)\Leftarrow Engage\ in\ diplomatic\ \emph{cooperation}^{-1}\left(x,\ z,\ t\right)$\\
  $\land\ \emph{p}_2\left(z,\ y,\ \ t\right)$   & $\land\ Express\ intent\ to\ settle \emph{dispute}\left(z,\ y,\ t\right)\ 0.67$\\
 \bottomrule
 \end{tabular}
 \caption{The cases of temporal rules mined from ICEWS05-15 with respective to each temporal rule pattern.}
 \label{table2}
 \end{table*}
 
To learn the time-independent representation of commonsense associated with events, the timestamp in each event is masked to convert the event quadruple $(s, p, o, t)$ into the factual triple $(s, p, o)$. Motivated by some typical commonsense KGs such as ConceptNet~\cite{conceptnet}, commonsense is represented as two concepts linked by a predicate. Therefore, we score each event in the view of commonsense via the learnable concept and predicate embeddings together with the proposed time-independent score function based on commonsense:
\begin{equation}
    E_2(s, p, o)=\mathrm{Re}\left(\sum_{i=1}^{k}{\left[\mathbf{s}_{c}\right]_{i}\cdot\left[\mathbf{p}_{c}\right]_{i}\cdot\left[{\bar{\mathbf{o}}}_{c}\right]_{i}}\right)
    \label{eq13}
\end{equation}
where $\mathbf{s}_c \in \mathbb{C}^k$, $\mathbf{p}_c \in \mathbb{C}^k$, $\mathbf{o}_c \in \mathbb{C}^k$ represent the concept embeddings in the $k$-dimension complex vector space with regard to subject $s$, predicate $p$ and object $o$, respectively. Particularly, $k$ should be set smaller than $d$ to enhance the abstract feature of entity concept embeddings.

\subsection{Optimization Objective}
\label{sec:3.5}

We employ the log-softmax loss function and N3 regularization to design the optimization objective for training:
\begin{align}\small
    L=&\sum_{(s,\ p,\ o,\ t)\in\mathcal{T}}\left(L_1+\lambda \cdot L_2\right) \label{eq14} \\
    L_1=&-\log\left(\frac{\exp\left(E_1\left(s,\ p,\ o,\ t\right)\right)}{\sum_{e_i\in\mathcal{E}}\exp\left(E_1\left(e_i,\ p,\ o,\ t\right)\right)}\right) \nonumber \\
    &-\log\left(\frac{\exp\left(E_1\left(s,\ p,\ o,\ t\right)\right)}{\sum_{e_i\in\mathcal{E}}\exp\left(E_1\left(s,\ p,\ e_i,\ t\right)\right)}\right) \label{eq15} \\
    &+\alpha_1 \left( \Vert \mathbf{s} \Vert_3^3 + \Vert \mathbf{p}_{t} \Vert_3^3 + \Vert \mathbf{p}_{r} \Vert_3^3 + \Vert \mathbf{o} \Vert_3^3 \right) \nonumber \\
    L_2=&-\log\left(\frac{\exp\left(E_2\left(s,\ p,\ o\right)\right)}{\sum_{e_i\in\mathcal{E}}\exp\left(E_2\left(e_i,\ p,\ o\right)\right)}\right) \nonumber \\
    &-\log\left(\frac{\exp\left(E_2\left(s,\ p,\ o\right)\right)}{\sum_{e_i\in\mathcal{E}}\exp\left(E_2\left(s,\ p,\ e_i\right)\right)}\right) \label{eq16} \\
    &+\alpha_2 \left( \Vert \mathbf{s}_c \Vert_3^3 + \Vert \mathbf{p}_{c} \Vert_3^3 + \Vert \mathbf{o}_c \Vert_3^3 \right) \nonumber
\end{align}
in which $L_1$ and $L_2$ represent the loss functions specific to the time-sensitive events and time-independent commonsense, respectively. $\mathcal{T}$ and $\mathcal{E}$ are the event set and the entity set in the TKG, respectively. $e_i$ denotes an entity in the entity set $\mathcal{E}$. $\alpha_1$ and $\alpha_2$ are defined as the N3 regularization weights corresponding to entity embeddings and predicate embeddings, respectively. $\lambda$ is the weight of commonsense representation in the overall loss function, which is applied for the trade-off between the time-sensitive and the time-independent representations of each event.

The whole optimization objective is the combination of the loss function in Eq.~\ref{eq14} and the predicate embedding regularization in Eqs.~\ref{eq7}-\ref{eq11}. Besides, our model is trained with the Adam optimizer~\cite{Adam} to learn the embeddings of entities, predicates, concepts and timestamps.

\section{Experiments and Results}

In this section, we introduce the experiment settings and exhibit some cases of temporal rules mined by our temporal rule learning algorithm. Then, the experimental results compared with some state-of-the-art baselines are provided to demonstrate the effectiveness of our model. Furthermore, the ablation study and case study are conducted.

\subsection{Experiment Settings}

 \noindent \textbf{Datasets.} Three commonly used datasets of TKGs are used in the experiments, namely ICEWS14~\cite{TA-TransE}, ICEWS05-15~\cite{TA-TransE} and Wikidata12k~\cite{HyTE}. Both ICEWS14 and ICEWS05-15 contain political events with specific timestamps. Wikidata12k~\cite{HyTE} is a subset of Wikidata~\cite{wikidata}, in which each time annotation is a timestamp or time interval. For each dataset, all the events are split into training, validation and test sets in a proportion of 80\%/10\%/10\% following some previous works~\cite{TComplEx, TERO}. Statistics for these datasets are presented in Table~\ref{table1}.
 
\begin{table*}[!t]\small
\renewcommand{\arraystretch}{1.5}
\centering
\renewcommand\tabcolsep{5pt}
\begin{tabular}{c|c|cccc|cccc|cccc}
\toprule
\multicolumn{2}{c|}{\multirow{2}*{Models}} & \multicolumn{4}{c|}{ICEWS14} & \multicolumn{4}{c|}{ICEWS05-15} & \multicolumn{4}{c}{Wikidata12k} \\
	\multicolumn{2}{c|}{}       & MRR	   & H@10	& H@3    & H@1	& MRR	& H@10	& H@3    & H@1	& MRR	& H@10	& H@3    & H@1\\
\midrule
& TransE    & 0.280	    & 0.637	&-	&0.094	&0.294	&0.663	&-	&0.090	&0.178	&0.339	&0.192	&0.100 \\
Typical & DistMult    & 0.439	&0.672	&-	&0.323	&0.456	&0.691	&-	&0.337	&0.222	&0.460	&0.238	&0.119 \\
Models & ComplEx    & 0.467	&0.716	&0.527	&0.347	&0.481	&0.729	&0.535	&0.362	&0.233	&0.436	&0.253	&0.123 \\
& RotatE    & 0.418	&0.690	&0.478	&0.291	&0.304	&0.595	&0.355	&0.164	&0.221	&0.461	&0.236	&0.116 \\
& QuatE    & 0.471	&0.712	&0.530	&0.353	&0.482	&0.727	&0.529	&0.370	&0.230	&0.416	&0.243	&0.125 \\
\midrule
& TTransE    & 0.255	&0.601	&-	&0.047	&0.271	&0.616	&-	&0.085	&0.172	&0.329	&0.185	&0.096 \\
&HyTE	&0.297	&0.655	&0.416	&0.108	&0.316	&0.681	&0.445	&0.116	&0.253	&0.483	&0.197	&0.147 \\
TKGE    &TeRo	&0.562	&0.732	&0.621	&0.468	&0.586	&0.795	&0.668	&0.469	&0.299	&0.507	&0.329	&0.198 \\
Models  &ATiSE	&0.550	&0.750	&0.629	&0.436	&0.519	&0.794	&0.606	&0.378	&0.252	&0.462	&0.288	&0.148 \\
&TComplEx	&0.610	&0.770	&0.660	&0.530	&0.660	&0.800	&0.710	&0.590	&0.331	&0.539	&0.357	&\underline{0.233} \\
&TeLM	&\underline{0.625}	&\underline{0.774}	&\underline{0.673}	&\underline{0.545}	&\underline{0.678}	&\underline{0.823}	&\underline{0.728}	&\underline{0.599}	&\underline{0.332}	&\underline{0.542}	&\underline{0.360}	&0.231 \\
\midrule
\textbf{Ours} & \textbf{LCGE}    &\textbf{0.667}	&\textbf{0.815}	&\textbf{0.714}	&\textbf{0.588}	&\textbf{0.730}	&\textbf{0.866}	&\textbf{0.776}	&\textbf{0.655}	&\textbf{0.429}	&\textbf{0.677}	&\textbf{0.495}	&\textbf{0.304} \\
\bottomrule
\multicolumn{2}{c|}{$\textbf{APG}$ $\uparrow(\%)$}	&\textbf{3.8}	&\textbf{4.1}	&\textbf{4.1}	&\textbf{4.3}	&\textbf{5.2}	&\textbf{4.3}	&\textbf{4.8}	&\textbf{5.6}	&\textbf{9.7}	&\textbf{13.5}	&\textbf{13.5}	&\textbf{7.3} \\
\multicolumn{2}{c|}{$\textbf{RPG} \uparrow(\%)$}	&\textbf{6.7}	&\textbf{5.3}	&\textbf{6.1}	&\textbf{7.9}	&\textbf{7.7}	&\textbf{5.2}	&\textbf{6.6}	&\textbf{9.3}	&\textbf{29.2}	&\textbf{24.9}	&\textbf{37.5}	&\textbf{31.6}\\
\bottomrule
\end{tabular}
\caption{Temporal KG completion results on three datasets. H1, H3 and H10 represent Hits@1, Hits@3 and Hits@10, respectively. \textbf{Bold} values are the best results and the second-to-best results are \underline{underlined} in all the models. APG and RPG indicate the absolute performance gains and the relative performance gains achieved by our model compared with the best-performing baseline TeLM. APG and RPG can be calculated by $APG=R_{ours} - R_{baseline}$ and $RPG=(R_{ours} - R_{baseline})/R_{baseline}$, where $R_{ours}$ and $R_{baseline}$ denote the results of our model and the baseline TeLM, respectively.}
\label{table3}
\end{table*}

 \noindent \textbf{Evaluation Protocol.} On the consideration that the amount of entities in any TKG is much more than that of predicates, predicting entities is more challenging than predicting predicates. Thus, the temporal KG completion task usually focuses on entity prediction such as $(s, p, ?, t)$ where the object is missing in this event. To achieve the prediction results, each entity $e_i$ in the TKG namely the candidate entity is filled in the object position to reconstruct a candidate event quadruple $(s, p, e_i, t)$. Then, the score for evaluating the plausibility of each candidate event is obtained from the time-sensitive and time-independent representations according to Eq.~\ref{eq12} and Eq.~\ref{eq13}, which is defined as
\begin{equation}
    E_{pred}(e_i)=E_1\left(s,p,e_i,t\right) + \lambda \cdot E_2\left(s,p,e_i\right)
    \label{eq18}
\end{equation}

\noindent \textbf{Baselines.} We select two types of state-of-the-art baseline models for comparison: 

(1) Some typical KGE models without time information, including TransE~\cite{Bordes:TransE}, DistMult~\cite{Distmult}, ComplEx~\cite{Trouillon:ComplEx}, RotatE~\cite{RotatE} and QuatE~\cite{QuatE}. 

(2) The previous well-performed TKGE models, including TTransE~\cite{TTransE}, HyTE~\cite{HyTE}, ATiSE~\cite{ATiSE}, TeRo~\cite{TERO}, TComplEx~\cite{TComplEx} and TeLM~\cite{TeLM}.

\noindent \textbf{Metrics and Implementation Details.}
We rank all the candidate events according to their scores calculated by Eq.~\ref{eq18} in descending order. Then, the rank of the correct event of the $i$-th test instance is defined as $rank_i$. We employ two commonly-used metrics for evaluating the results of TKGC:

(1) The reciprocal mean rank (MRR) of the correct events, which can be calculated by
\begin{equation}
    MRR = \frac{1}{N} \cdot \sum_{i}^{N} \frac{1}{rank_i}
\end{equation}
where $N$ is the total amount of test instances.

(2) The ratio of the correct events ranked in the top n (Hits@n), which is calculated by
\begin{equation}
    Hits@n = \frac{1}{N} \cdot \sum_{i}^{N} \mathbb{I}(rank_i \leq n)
\end{equation}
where the value of function $\mathbb{I}(rank_i \leq n)$ is 1 if $rank_i \leq n$ is true. $n$ is usually set as 1, 3 or 10. Note that the higher MRR and Hits@n indicate better performance.

In particular, our model represents the timeliness of each event with its timestamp, it is necessary to convert an event with a time interval in wikidata12k into two events along with the timestamps at both ends of the time interval. In the inference stage, the score of each event with a time interval in wikidata12k is achieved by averaging the individual scores of two events with endpoint timestamps of the time interval. Besides, the open-source rule learning tool AMIE+~\cite{Galarrage:AMIE} is utilized to mine the static rules for its convenience and good performance.

\begin{table*}[!t]\small
\renewcommand{\arraystretch}{1.1}
\centering
\renewcommand\tabcolsep{7pt}
\begin{tabular}{c|cccc|cccc}
\toprule
\multirow{2}*{Models} & \multicolumn{4}{c|}{ICEWS14} & \multicolumn{4}{c}{ICEWS05-15} \\
	& MRR	& Hits@10	& Hits@3  & Hits@1	& MRR	& Hits@10	& Hits@3  & Hits@1\\
\midrule
LCGE        &\textbf{0.667}	&\textbf{0.815}	&\textbf{0.714}	&\textbf{0.588}	&\textbf{0.730}	&\textbf{0.866}	&\textbf{0.776}	&\textbf{0.655} \\
\midrule
-RGPR        &0.663	&0.812	&0.711	&0.585	&0.720	&0.863	&0.770	&0.645 \\
-TIS      &0.621	&0.774	&0.664	&0.540	&0.679	&0.826	&0.728	&0.598 \\
\bottomrule
\end{tabular}
\caption{Ablation study of eliminating RGPR module and time-independent score from the whole model LCGE.}
\label{table4}
\end{table*}

We conduct all the experiments in Pytorch and on a GeForce GTX 2080Ti GPU. The batch size is set as 1024. The thresholds of SC and HC in our temporal rule learning algorithm are both fixed to 0.1 on all the datasets. We tune the hyper-parameters by grid search on the validation sets. Specifically, the dimension of embeddings $d$ is selected in the range of $\{100, 200, 500, 1000, 2000\}$. The dimension of embeddings specific to commonsense $k$ is set as the half of $d$. The learning rate $l_r$ is tuned in $\{0.01, 0.05, 0.1, 0.5\}$, the weights of N3 regularization $\alpha_1$ and $\alpha_2$ are chosen in $\{0.005, 0.01, 0.05, 0.1, 1.0\}$, the weight of commonsense in the whole loss function $\lambda$ is adjusted in $\{0.1, 0.5, 1.0\}$.

\subsection{Cases of Temporal Rules}

Some cases of temporal rules mined by our temporal rule learning algorithm from ICEWS05-15 are exhibited in Table~\ref{table2}. We observe that each temporal rule has a confidence level, which corresponds to the weight of the rule-guided predicate embedding regularization derived from this rule. These temporal rules are friendly for humans to understand, which benefits the explainability of the predicted results.

\subsection{Experimental Results}

The comparison results$\footnote{We have corrected the mistake in the previous version of our code and updated the results in this revision.}$ between our LCGE model with both the traditional KGE models and the existing TKGE models are provided in Table~\ref{table3}. The results of all the baseline models on ICEWS14 and ICEWS05-15 are directly taken from the original paper of TeLM~\cite{TeLM}, and the experimental results of all baselines on Wikidata12k are obtained by running the source code corresponding to each baseline model. We observe that \textbf{our proposed LCGE model outperforms all the baselines consistently and significantly on all the datasets}. Particularly, compared with the best-performing baseline model TeLM, our approach obtains substantial performance improvements. Specifically, in terms of the absolute performance gains (APG), our model achieves \textbf{3.8}\%/\textbf{5.2}\%/\textbf{9.7}\% as to MRR and \textbf{4.3}\%/\textbf{5.6}\%/\textbf{7.3}\% as to Hits@1 on ICEWS14/ICEWS05-15/Wikidata12k. With respect to the relative performance gains (RPG), our model achieves \textbf{6.7}\%/\textbf{7.7}\%/\textbf{29.2}\% as to MRR and \textbf{7.9}\%/\textbf{9.3}\%/\textbf{31.6}\% as to Hits@1 on ICEWS14/ICEWS05-15/Wikidata12k.

In particular, the recent TKGE baselines and our model LCGE all outperform the traditional KGE models, which emphasizes the significance of learning the time-sensitive representation of events for TKGC task. Furthermore, the more superior performance of our model compared with the TKGC baselines demonstrates the effectiveness of evaluating the plausibility of each event from the perspectives of the better time-sensitive representation and the auxiliary time-independent commonsense.

\subsection{Ablation Study}

To verify the effectiveness of each module in our scheme, we observe the performances of two ablated models: (1) eliminating the RGPR module from the whole model (-RGPR), and (2) removing the time-independent score (-TIS) during both training and inference stages. The results of the ablation study as shown in Table~\ref{table4} demonstrate that eliminating RGPR or TIS both has an obvious impact on the performance of the whole model. Besides, omitting TIS shows more performance drop, which suggests the vital role of predicting events in the view of commonsense. Furthermore, the number of high-quality temporal rules is limited, leading to relatively less performance improvement derived from temporal rules. The result on Wikidata12k is not presented since the number of high-quality temporal rules mined from Wikidata12k is too small to verify the effectiveness of the ablated model -RGPR. Thus, how to mine more temporal rules would be studied in-depth in future research.

\subsection{Case Study}

\begin{figure}
    \centering
    \includegraphics[scale=0.45]{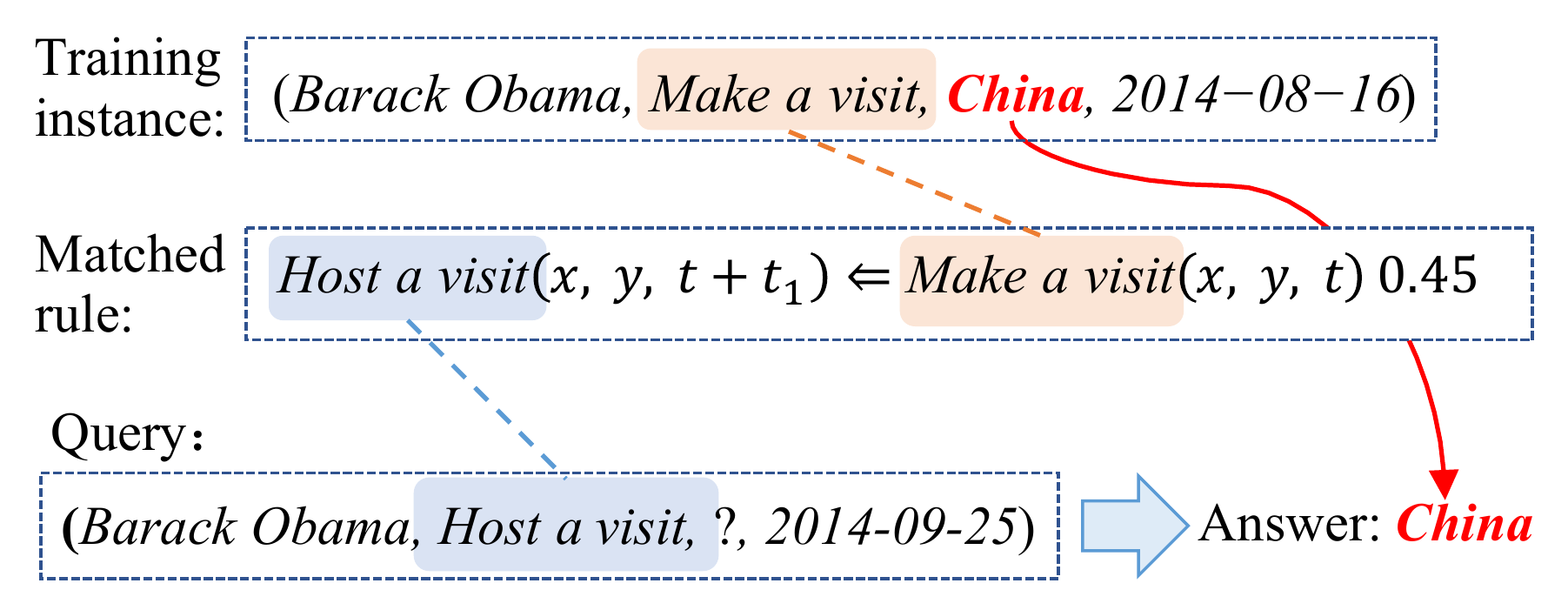}
    \caption{A case study of TKGC with explainability via the causality among events with a temporal rule.}
    \label{fig:case}
\end{figure}

It is noteworthy that our approach could employ the symbolic temporal rules to declare the inference process explicitly and enhance the explainability that all the existing TKGE models are lack of. On account of a TKGC query $(China, Make$ $a\ Visit,\ ?,\ $2012-11-16$)$ as shown in Figure~\ref{fig:case}, our model predicts the entity $South\ Korea$ with the highest score. More particularly, a temporal rule $Host\ a\ visit(x,\ y,\ t+t_1)\ \Leftarrow Make\ a\ visit (x, y, t)$ could be matched to the given query. Meanwhile, an event $(China,\ Host\ a\ visit,\ South\ Korea,$ 2012-11-20$)$ in the training set shares the same subject with the query, which is a grounding of the head atom of the matched rule. Therefore, $South\ Korea$ could be directly deduced by the temporal rule, which is consistent with the result obtained by our model and facilitates the explainability of the result.

\section{Conclusion}

In this paper, we propose a novel and effective logic and commonsense-guided embedding model LCGE for TKGC task. As far as we can be concerned, our model is the first to learn both the time-sensitive representation of events involving timeliness and causality as well as the time-independent commonsense representation, simultaneously. Specifically, the causality among events could be learned via our temporal rule learning algorithm and the temporal rule-guided predicate embedding regularization. The experimental results of the TKGC task on three datasets illustrate the significant performance improvements obtained by our model LCGE compared with the state-of-the-art baselines. More interestingly, our model could provide the explainability of results according to causality inference via temporal rules.

\section*{Acknowledgements}
\label{acknowledgement}

This work was partially supported by Zhejiang Science and Technology Plan Project (No. 2022C01082), the National Natural Science Foundation of China (No. 62072022, 61772054) and the Fundamental Research Funds for the Central Universities.

\bibliography{aaai23}

\appendix

\section{Appendix}
\label{sec:appendix}

\subsection{Description of the Temporal Rule Patterns}
\label{appx1}
 
(1) $p_2(x, y, t+t_1) \Leftarrow p_1(x, y, t)$. With regard to a length-1 rule, the body atom $p_1(x, y, t)$ and the head atom $p_2(x, y, t+t_1)$ occur at different times. This temporal rule pattern signifies that two events containing predicates $p_1$ and $p_2$ would occur one after the other if both the subject and object are the same, respectively. For instance, given a temporal rule $Consult\left(x,\ y,\ t+t_1\right)\ \Leftarrow\ Express\ intent\ to\ meet \ or$ $\ negotiate \left(x,\ y,\ t\right)$ and an event $(China, Express\ intent$ $\ to\ meet$ $\ or\ negotiate, U.S.A., TS_1)$ occurring at the timestamp $TS_1$, we could easily deduce that an event $(China,$ $Consult,$ $U.S.A., TS_1+T)$ will happen after $TS_1$.

(2) $p_2(x, y, t) \Leftarrow p_1(x, y, t)$. For a length-1 rule, the body atom $p_1(x, y, t)$ and the head atom $p_2(x, y, t)$ occur simultaneously. This pattern is equivalent to the length-1 static rule.

(3) $p_3(x, y, t+t_1+t_2) \Leftarrow p_1(x, z, t) \wedge p_2(z, y, t+t_1)$. The three atoms in this pattern of temporal rule occur at different time from each other. This pattern implies the three causally related events occur one after the other. For instance, the temporal rule $Express\ intent\ to\ meet\ or\ negotiate(x,\ y,\ t+t_1+t_2)$ $\Leftarrow Engage\ in\ negotiation^{-1}\left(x,\ z,\ t\right) \wedge Express\ intent\ to$ $\ cooperate\left(z,\ y,\ \ t+t_1\right)$ belongs to this temporal rule pattern, where the predicate $Engage\ in\ negotiation^{-1}$ is the inverse version of the predicate $Engage\ in\ negotiation$.

(4) $p_3(x, y, t+t_1) \Leftarrow p_1(x, z, t) \wedge p_2(z, y, t)$. Specific to this rule pattern, the two events corresponding to the body atoms $p_1(x, z, t)$ and $p_2(z, y, t)$ such as $(s, p_1, e, t)$ and $(e, p_2, o, t)$ occur at the timestamp $t$, and then the event $(s, p_3, o, t+t_1)$ associated with the atom $p_3(x, y, t+t_1)$ would happen.

(5) $p_3(x, y, t) \Leftarrow p_1(x, z, t) \wedge p_2(z, y, t)$. Similar to the pattern (2), this pattern of the temporal rule indicates that all the events corresponding to the atoms occur at the same time, which is equivalent to the length-2 static rule.

\subsection{Evaluation Criteria of Length-2 Temporal Rules}

For the length-2 candidate temporal rule in the general form of $p_3\left(x,\ y,t+T_1+T_2\right)\Leftarrow p_1\left(x,\ z,\ t\right)\land p_2\left(z,\ y,t+T_1\right)$, there are three diverse patterns with the different values of $T_1$ and $T_2$. In allusion to each timestamp $t$, the SD, SC and HC are defined as:
\begin{align}\small
    &SD= \nonumber\\
    &\begin{cases}
    \sum_{t_{1i}\in[1,\ W_T]}{\sum_{t_{2i}\in[1,\ W_T]}{\#(a,c): p_1\left(a,b,t\right)}}\\
    \quad \land\ p_2\left(b,c,t+t_{1i}\right)\land p_3\left(a,c,t+T_{1i}+T_{2i}\right),\\
    \qquad \qquad \qquad \qquad \qquad \qquad \ \ \ T_1>0,T_2>0\\
     \sum_{t_{2i}\in[1,\ W_T]}{\#\left(a,c\right):p_1\left(a,b,t\right)\land p_2\left(b,c,t\right)}\\
     \quad \land\ p_3\left(a,c,t+t_{2i}\right),\qquad \ \ \ \ \ \ \  T_1=0,T_2>0 \tag{18}\\
    \#\left(a,c\right):p_1\left(a,b,t\right)\land p_2\left(b,c,t\right) \land p_3\left(a,c,t\right),\\ \qquad \qquad \qquad \qquad \qquad \qquad \ \ \ \ \ \ T_1=T_2=0
    \label{eq4}
    \end{cases}
    \end{align}
    \begin{align}
    &SC= \nonumber \\
    &\begin{cases}
    \frac{1}{\left|W_T\right|}\cdot\frac{SD}{\sum_{t_{1i}\in[1,\ W_T]}{\#\left(a,c\right):p_1\left(a,b,t\right)\land p_2\left(b,c,t+t_{1i}\right)}}, \tag{19} \\
    \qquad \qquad \qquad \qquad \qquad \qquad \qquad \quad \ T_1>0  \label{eq5} \\
    \frac{SD}{\#\left(a,c\right):p_1\left(a,b,t\right)\land p_2\left(b,c,t\right)}, \qquad \qquad \ \ \ \ \ \ T_1=0
    \end{cases}\\
    &HC=
    \begin{cases}
    \
    \frac{SD}{\sum_{t_{1i}\in\left[{1,W}_T\right]}\sum_{t_{2i}\in\left[{1,W}_T\right]}{\#\left(a,c\right):p_3\left(a,c,t+t_{1i}+t_{2i}\right)}}, \tag{20} \\
    \qquad \qquad \qquad \qquad \qquad \quad \ \ \ \ \ \ T_1>0, T_2>0 \\
    \frac{SD}{\sum_{t_{1i}\in\left[{1,W}_T\right]}{\#\left(a,c\right):p_3\left(a,c,t+t_{1i}+t_{2i}\right)}}, \\
    \qquad \qquad \qquad \qquad \qquad \quad \ \ \ \ \ \ T_1=0, T_2>0 \\
    \frac{SD}{\#\left(a,c\right):p_3\left(a,c,t\right)}, \qquad \qquad \quad \quad \ \ \ \ \ T_1=T_2=0  \label{eq6}
    \end{cases}
\end{align}

It is noteworthy that the temporal rule with the settings (1) $T_1>0, T_2>0$, (2) $T_1>0, T_2=0$, (3) $T_1=T_2=0$ correspond to the temporal rule patterns (3), (4) and (5).

\subsection{The Algorithm of Our Proposed Temporal Rule Learning Module}

For better understanding, Algorithm \ref{alg1} summarizes the whole procedure of our temporal rule learning procedure.
\begin{algorithm}  
  \caption{our temporal rule learning module.}  
  \label{alg1}
  \noindent \KwIn{
$\mathcal{G}_s$, $\mathcal{G}_t$: the GSKG and the TKG \newline
$T$, $W_T$: the number of timestamps and the size of time window \newline
$T_{sc}$, $T_{hc}$: the thresholds of SC and HC of temporal rules
}

\KwOut{
$\mathcal{R}_t$: the set of temporal rules satisfying $T_{sc}$ and $T_{hc}$ \newline
}  
\textbf{Mine} static rules by a rule mining tool such as AMIE+ from $\mathcal{G}_s$\;
\textbf{Convert} each static rule into candidate temporal rules via temporal rule patterns\;
\While{select a candidate temporal rule $Ru$}{
        \For{t=1,2,\dots, T}{
          \textbf{Generate} two sub-graphs $\mathcal{G}_{s1}$ and $\mathcal{G}_{s2}$ via fusing the events in the time intervals [$t$+1, $t$+$W_T$] and [$t$+$W_T$+1, $t$+2$W_T$] from $\mathcal{G}_t$\;
          \If {the length of $Ru$ is 1}
          {
          \textbf{Calculate} the SC and HC of $Ru$ by Eqs.~2-4\;
          }
          \Else
          {
            \textbf{Calculate} the SC and HC of $Ru$ by Eqs.~\ref{eq4}-\ref{eq6}\;
          }
        }
      \If{the mean value of SC and HC at all the timestamps are both higher than $T_{sc}$ and $T_{hc}$}{
      Add $Ru$ into $\mathcal{R}_t$\;
      }
  }
\end{algorithm}

\subsection{Proof of Representing Causality Among Events via RGPR Mechanism}

In this paper, we take the temporal rule patterns (1) and (3) as examples, and provide the lemma proofs that the temporal rule-guided predicate embedding regularization $\mathbf{G}$ could represent the causality among events.

\noindent \textbf{Lemma 1.} On account of the temporal rule pattern (1) $p_2(x, y, t+t_1) \Leftarrow p_1(x, y, t)$, the temporal rule-guided predicate embedding regularization $\mathbf{G}$ specific to Eq.~5 could represent the causality between two events.
\begin{proof}
Given an event $(s, p_1, o, t)$, we could obtain the causality term of the time-sensitive score specific to this event transformed at the timestamp $t+t_1$ according to Eq.~10 and the time transfer operator $\mathbf{T}$, which can be written as
\begin{equation}
    \mathrm{Re}\left(\mathbf{s}^\top\ diag(\mathbf{T} \circ \mathbf{p}_{r1})\ \bar{\mathbf{o}}\right) = 1 \tag{21}\label{eq19}
\end{equation}

According to $\mathbf{G}$ defined in Eq.~5, the association between $\mathbf{p}_{r1}$ and $\mathbf{p}_{r2}$ satisfies:
\begin{equation}
    \mathbf{T}\circ \mathbf{p}_{r1}=\mathbf{p}_{r2} \tag{22}\label{eq20}
\end{equation}
Substituting Eq.~22 into Eq.~21, we can derive that
\begin{equation}
    \mathrm{Re}\left(\mathbf{s}^\top\ diag(\mathbf{p}_{r2})\ \bar{\mathbf{o}}\right) = 1 \tag{23}\label{eq21}
\end{equation}

It can be discovered that the event $(s, p_2, o, t+t_1)$ holds if $(s, p_1, o, t)$ occurs. Therefore, the causality between two events occurring at different timestamps could be represented via the temporal rule-guided predicate embedding regularization $\mathbf{G}$ of the pattern (1) defined in Eq.~\ref{eq7}.
\end{proof}

\noindent \textbf{Lemma 2.} For the temporal rule pattern $p_3(x, y, t+t_1+t_2) \Leftarrow p_1(x, z, t) \wedge p_2(z, y, t+t_1)$, the temporal rule-guided predicate embedding regularization in Eq.~\ref{eq9} can represent the causality among three events at different timestamps.

\begin{proof}
If there are two events $(s, p_1, e, t)$ and $(e, p_2, o, t+t_1)$ holds, the causality term of the time-sensitive scores specific to these two events transformed to the timestamp $t+t_1+t_2$ are achieved according to Eq.~10 as follows:
\begin{align}
    \mathrm{Re}\left(\mathbf{s}^\top\ diag(\mathbf{T} \circ \mathbf{T} \circ \mathbf{p}_{r1})\ \bar{\mathbf{e}}\right) &= 1 \tag{24}\label{eq22}\\
    \mathrm{Re}\left(\mathbf{e}^\top\ diag(\mathbf{T} \circ \mathbf{p}_{r2}) \bar{\mathbf{o}}\right) &= 1 \tag{25}\label{eq23}
\end{align}
In terms of the following equation
\begin{align}
   &\mathbf{s}^\top\ diag(\mathbf{T} \circ \mathbf{T} \circ \mathbf{p}_{r1})\ \bar{\mathbf{e}} \mathbf{e}^\top\ diag(\mathbf{T} \circ \mathbf{p}_{r2}) \bar{\mathbf{o}} \nonumber \\
   &=\mathbf{s}^\top\ diag(\mathbf{T} \circ \mathbf{T} \circ \mathbf{p}_{r1} \circ \mathbf{T} \circ \mathbf{p}_{r2})\ \bar{\mathbf{o}} \tag{26}\label{eq24}
\end{align}
and Eqs.~24 and 25, we could derive that
\begin{align}
    &\mathrm{Re}\left(\mathbf{s}^\top\ diag(\mathbf{T} \circ \mathbf{T} \circ \mathbf{p}_{r1} \circ \mathbf{T} \circ \mathbf{p}_{r2})\ \bar{\mathbf{o}}\right) = 1-  \tag{27} \label{eq25}\\ &\mathrm{Im}\left(\mathbf{s}^\top\ diag(\mathbf{T} \circ \mathbf{T} \circ \mathbf{p}_{r1}) \bar{\mathbf{e}}\right) \cdot \mathrm{Im}(\mathbf{e}^\top\ diag(\mathbf{T} \circ \mathbf{p}_{r2}) \bar{\mathbf{o}}) \nonumber
\end{align}
Furthermore, if the following equation 
\begin{equation}
    \mathrm{Im}\left(\mathbf{s}^\top\ diag(\mathbf{T} \circ \mathbf{T} \circ \mathbf{p}_{r1}) \bar{\mathbf{e}}\right) \cdot \mathrm{Im}(\mathbf{e}^\top\ diag(\mathbf{T} \circ \mathbf{p}_{r2}) \bar{\mathbf{o}})=0 \tag{28}
\end{equation}
is satisfied, Eq. 27 can be rewritten as:
\begin{equation}
    \mathrm{Re}\left(\mathbf{s}^\top\ diag(\mathbf{T} \circ \mathbf{T} \circ \mathbf{p}_{r1} \circ \mathbf{T} \circ \mathbf{p}_{r2})\ \bar{\mathbf{o}}\right) = 1 \tag{29} \label{eq26}
\end{equation}
Substituting $\mathbf{G}$ defined in Eq.~7 into Eq.~29, we can achieve
\begin{equation}
    \mathrm{Re}\left(\mathbf{s}^\top\ diag(\mathbf{p}_{r3})\ \bar{\mathbf{o}}\right) = 1 \tag{30} \label{eq27}
\end{equation}
Therefore, we prove that the event $(s, p_3, o, t+t_1+t_2)$ is valid if $(s, p_1, e, t)$ and $(e, p_2, o, t+t_1)$ successively occur via the temporal rule-guided predicate embedding regularization $\mathbf{G}$ of the pattern (3), which indicates the causality among three events occurring at different timestamps.
\end{proof}

The proofs of the causality among events corresponding to the other three temporal rule patterns modeled by our designed temporal rule-guided predicate embedding regularization $\mathbf{G}$ can be obtained similarly to the above proofs.

From the proofs corresponding to Lemma 1 and Lemma 2, the causality term in the time-sensitive score of a candidate event causally related to the other already occurred events would be a higher value. More interestingly, it can explain the effectiveness of representing the causality of events via our temporal rule-guided predicate embedding regularization mechanism and facilitate higher accuracy of TKGC.

\end{document}